\documentclass[onefignum,onetabnum]{siamart220329}

\usepackage{lipsum}
\usepackage{amsfonts}
\usepackage{graphicx}
\usepackage{epstopdf}
\usepackage{multirow}
\usepackage{algorithmic}

\ifpdf
  \DeclareGraphicsExtensions{.eps,.pdf,.png,.jpg}
\else
  \DeclareGraphicsExtensions{.eps}
\fi

\definecolor{red2}{RGB}{204,0,0}
\definecolor{blue2}{RGB}{0,103,165}
\hypersetup{colorlinks,linkcolor=[RGB]{0,103,165},citecolor=[RGB]{180,0,0}}
\usepackage[nocompress]{cite}

\newsiamremark{remark}{Remark}
\newsiamremark{hypothesis}{Hypothesis}
\crefname{hypothesis}{Hypothesis}{Hypotheses}
\newsiamthm{claim}{Claim}

\def\be{\begin{equation}}
\def\ee{\end{equation}}

\def\x{\mathbf{x}}

\def\xh{\widehat{\x}}

\def\y{\mathbf{y}}

\def\f{\mathbf{f}}
\def\g{\mathbf{g}}
\def\z{\mathbf{z}}

\def\Z{\mathbf{Z}}
\def\Y{\mathbf{Y}}

\def\Rs{\mathbb{R}}

\def\G{\mathbf{G}}

\def\Gh{\mathbf{\widehat{G}}}
\def\N{\mathbf{N}}

\def\T{\mathbf{T}}
\def\Pi{\mathbf{\Phi}}
\def\PPh{\mathbf{\Phi}}

\def\D{\mathbf{D}}

\headers{Modelling of Multiscale SDE}{Yuan Chen and Dongbin Xiu}

\title{Data-driven Effective Modeling of Multiscale Stochastic Dynamical Systems}

\author{Yuan Chen\and Dongbin Xiu\thanks{E-mail addresses: \texttt{\{chen.11050, xiu.16\}@osu.edu}. Department of Mathematics, The Ohio State University, Columbus, OH 43210, USA. Funding: This work was partially supported by AFOSR FA9550-22-1-0011.}}

\usepackage{amsopn}

\ifpdf
\hypersetup{
  pdftitle={Title},
  pdfauthor={Yuan Chen}
}
\fi

\begin{document}

\maketitle

\begin{abstract}
We present a numerical method for learning the dynamics of slow components of unknown multiscale stochastic dynamical systems. While the governing equations of the systems are unknown, bursts of observation data of the slow variables are available. By utilizing the observation data, our proposed method is capable of constructing a generative stochastic model that can accurately capture the effective dynamics of the slow variables in distribution. We present a comprehensive set of numerical examples to demonstrate the performance of the proposed method.
\end{abstract}



\section{Introduction}
Many mathematical models are established based on complex dynamics with multiple time scales. Understanding and simulating such systems are challenging due to the separation of the scales. 
One common approach is to establish effective equations, also known as averaged equations, for the slow processes by averaging out the fast variables with respect to their invariant measure, cf. \cite{pavliotis2008multiscale, khasminskij1968principle}. Another popular way resorts to identifying and modeling certain low-dimensional quantities to represent the dynamics, for example, invariant manifolds \cite{duan2003invariant,schmalfuss2008invariant}, transition pathways \cite{weinan2004metastability}, reaction coordinates \cite{coifman2008diffusion}, etc. More recently, efforts have been made to directly model the observables using data driven approaches. See, for example, \cite{dsilva2016data, feng2023learning, ye2024nonlinear, champion2019discovery, zielinski2022discovery}.

The focus of this paper is on numerical modeling of the effective dynamics of unknown slow-fast multiscale stochastic dynamical systems. More specifically, we assume that observation data of the slow variables are available, while the governing equations for the systems may not. The goal is to construct a numerical model that can accurately capture the slow dynamics, which in turn enables system analysis and prediction of the stochastic dynamics. The proposed method falls into a broader area of discovery of dynamics from observational data, which has attracted growing attention in recent years. Most of the methods focus on deterministic systems, cf. \cite{raissi2019physics, raissi2018multistep, brunton2016discovering, li2020fourier, owhadi2021computational, qin2019data, Churchill_2023}. For stochastic systems, the presence of noises poses new challenges to the design of effective data-driven methods. Several methods have been developed by using Gaussian processes (\cite{yildiz2018learning, pmlr-v1-archambeau07a, darcy2022one, opper2019variational}), or deep neural networks (DNNs) \cite{chen2023data, yang2022generative, chen2021solving, zhang2022multiauto, dietrich2023learning}, etc. More recently, a framework of stochastic flow map learning ( \cite{chen2023learning}) was proposed. By using generative machine learning models, the approach was shown to be highly effective in modeling unknown stochastic dynamical systems.

The framework of sFML is an extension of  FML (flow map learning, cf. \cite{qin2019data}) for modeling deterministic systems. The FML approach utilizes data to learn the flow map between state variables recorded on consecutive time steps. Originally developed to learn unknown autonomous dynamical systems, it was extended to learning nonautonomous dynamical systems \cite{qin2021data}, parametric systems \cite{QinChenJakemanXiu_IJUQ}, PDEs \cite{chen2022deep, wu2020data}, as well as systems when only a subset of state variables are observed \cite{FuChangXiu_JMLMC20}.  For unknown stochastic systems, the sFML approach incorporates a generative model in the learned flow map operator, such that the operator becomes stochastic and a weak approximation (in distribution) to the true stochastic flow map. Different generative models can be incorporated, for example, Generative Adversarial Networks (GANs) \cite{chen2023data}, auto-encoders \cite{xu2023learning}, and normalizing flows \cite{chen2024modeling}.

In this work, we adopt the sFML approach for learning the dynamics of the learning slow variables in multiscale stochastic dynamical systems. Once successfully constructed, the sFML model generates stochastic trajectories of the slow state variables for arbitrarily given initial conditions. The learned sFML model is a weak approximation (in distribution) to the true and unknown slow dynamics. Consequently, our model serves as an effective model for the slow dynamics, constructed by data of the slow variables but not the fast variables. Unlike \cite{FuChangXiu_JMLMC20,CiCP-25-947} for deterministic dynamical systems with missing variables, our sFML model does not require memory terms. This owes to the distinctive feature of multiscale stochastic dynamics \textemdash~ fast variables are quickly incorporated into ("slaved to") the dynamics of the slow variables, in the form of statistics, which independently contributes to the distribution of slow variables. The 
reduced system formed by only the slow variables thus becomes approximately Markovain. The impact of the missing dynamics of the fast variable is diminishingly small and presents itself, at best, as noises.
For the generative model of the sFML model,
we utilize conditional normalizing flow (cf. \cite{papamakarios2021normalizing}) model to approximate the stochasticity of the slow system.  
Other kind of generative models can also be considered in the sFML model, as shown the \cite{chen2023learning, xu2023learning, chen2024modeling}.
%

\section{Setup}
Let $\Omega$ be an event space and $T$ be a finite time horizon, we consider a $d$-dimensional stochastic system $\mathbf{u}^\varepsilon:\Omega\times [0,T] \mapsto \mathbb{R}^d$. The state variables $\mathbf{u}$ can be separated into slow variables $\x \in \mathbb{R}^{\ell}$ and fast variable $\y \in \mathbb{R}^{d-\ell}$, and are driven by an unknown stochastic differential equations:
\be
\label{full}
\left\{
\begin{split}
    d\x_t^{\varepsilon}&=\mathbf{f}\left(\x_t^{\varepsilon},     \y_t^{\varepsilon}\right)+\boldsymbol{\sigma}(\x_t^{\varepsilon}, \y_t^{\varepsilon})d\mathbf{W}_t^1, \\ 
    d\y_t^{\varepsilon}&=\frac{1}{\varepsilon} \mathbf{g}\left(\x_t^{\varepsilon}, \y_t^{\varepsilon}\right)+\frac{1}{\sqrt{\varepsilon}}\boldsymbol{\beta}\left(\x_t^{\varepsilon}, \y_t^{\varepsilon}\right)d\mathbf{W}_t^2,
\end{split}
\right.
\ee
where $\mathbf{f}$ and $\boldsymbol{\sigma}$, $\mathbf{g}$ and $\boldsymbol{\beta}$ are drift and diffusion functions, $\mathbf{W}_t^i$ are (independent) $m_i$-dimensional Brownian motions, $m_1+m_2 \geq 1$. The parameter $0<\varepsilon\ll 1$ measures the separation of time-scale in the system. We emphasize that the form of equations as well as the Brownian motion noises are unknown.


\subsection{Objective}

Although \eqref{full} is \emph{unknown}, in the sense that the functions $\mathbf{f}$, $\boldsymbol{\sigma}$, $\mathbf{g}$ and $\boldsymbol{\beta}$ are  \emph{unknown}, we assume that historical data of the slow variable $\x_t^\varepsilon$ are available. More specifically, let $t_0<t_1<...$ be discrete time instants. We observe $N_T\geq 1$ pieces of time history data of the slow variables $\x_t^\varepsilon$:
\be
\label{sqdata}\left(\x^\varepsilon(t_0^{(i)}),\x^\varepsilon(t_1^{(i)}),...,\x^\varepsilon(t_{L_i}^{(i)}) \right), \qquad i=1,2,...
\ee
where for simplicity we assume a constant time lag $\Delta = t_{i}-t_{i-1} \sim \mathcal{O}(1)$ and the length of all pieces of data is the same, $L_i = L$, $\forall i$. We remark that data of the fast variables are not required in this work.

Our goal is to construct a numerical model involving \emph{only} the slow variables $\x^\varepsilon_t$ for the dynamics of the slow variables of the multiscale system \eqref{full}, without the involvement of the fast variables $\y^\varepsilon_t$. The model predictions $\xh$ shall be an approximation, in distribution, to the true (and unknown) slow dynamics, i.e., 
\be
    \xh(t_i;\x_0) \stackrel{d}{\approx} \x^\varepsilon(t_i;\x_0),\qquad i=1,2,...
\ee
for arbitrarily given initial condition
$\x_0$.

\subsection{Related Work}
The current work is an extension of the FML (flow map learning) methodology for modeling deterministic unknown dynamical systems and its stochastic extension sFML for modeling unknown stochastic dynamical systems. 

For an unknown deterministic autonomous system,
$\frac{d\x}{dt} = \f(\x)$, $\x\in\mathbb{R}^d$,
where $\f$ is unknown. The FML framework seeks to approximate the unknown flow map
$ \x_n = \PPh_{t_n-t_s}(\x_s)$ by using historical trajectory data. Specifically, FML method fits a model
$$
\x_{n+1} = \widetilde{\PPh}_{\Delta t}(\x_n), 
$$
where $\widetilde{\PPh}_{\Delta t} \approx  {\PPh}_{\Delta t}$ is a
numerical approximation of the true flow map. Once trained, the FML model can be used as a time-marching scheme
to predict the system response under new initial conditions.
This framework was
proposed in  \cite{qin2019data}, with extensions to parametric systems
\cite{QinChenJakemanXiu_IJUQ}, partially observed dynamical systems \cite{FuChangXiu_JMLMC20}, as well as non-autonomous deterministic
system \cite{qin2021data}.

For learning stochastic dynamics,
$
\frac{d\x}{dt} = \f(\x,\omega(t)),
$
where $\omega(t)$ represents an unknown stochastic process driving the system. The work of
\cite{chen2023learning} developed stochastic flow map learning
(sFML). Assuming the system satisfies time-homogeneous property 
$
\mathbb{P}(\mathbf{x}_{s+\Delta
  t}|\mathbf{x}_{s})=\mathbb{P}(\mathbf{x}_{\Delta
  t}|\mathbf{x}_{0})$, $s\geq 0$ (c.f. \cite{oksendal2003stochastic}),
the method uses observation data on the state variable $\x$ to
construct a one-step generative model
$$
    \x _{n+1}=\G_{\Delta t}(\x_n; \z),
    $$
    where $\z$ is a random variable with known distribution (e.g.,
    standard Gaussian). The function $\G$, termed stochastic flow map,
    approximates the conditional distribution
$
\G_{\Delta t}(\mathbf{x}_{s}; \z) \sim
\mathbb{P}(\mathbf{x}_{s+\Delta t}|\mathbf{x}_{s}).
$
Subsequently, the sFML model becomes a weak approximation, in
distribution, to the true stochastic dynamics. Different generative
models can be employed under the sFML framework, e.g., generative adversarial networks (GANs) \cite{chen2023learning}, autoencoder \cite{xu2023learning}, etc. Recently, it was extended to stochastic systems subjected to external excitation  using normalizing flow \cite{chen2024modeling}.

\subsection{Contribution}
The contribution of the current work is on the construction of data-driven effective model for the slow variables of the multiscale stochastic system \eqref{full}. 
Under the assumption that the fast variables quickly reach their stationary measure and become "slaved" to the slow variables, we develop a sFML model that involves only the slow variables for accurate predictions of the effective dynamics of the original system. Such a modeling approach is highly useful, as in practice one often only has observation data on the slow variables because the fast variables are difficult, if not impossible, to observe. The proposed work can be considered as an extension of the sFML modeling methodology from general stochastic system to multiscale stochastic system.
\section{Method Description}
In this section, we describe the proposed method in detail.

\subsection{Mathematical Motivation}

\begin{figure}[htbp]
  \centering
  \includegraphics[width=.98\textwidth]{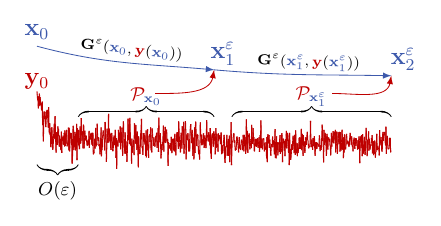}
  \caption{Illustration of how the fast variables $\y$ quickly become slaved to the slow variables $\x$ in multiscale system.}
  \label{fig:multiscale}
\end{figure}

Let us consider the unknown full system \eqref{full} over a time interval $[t_n, t_{n+1}]$, $n\geq 0$,
\be \label{x_n}
\x_{n+1}^\varepsilon = \x_n^\varepsilon + \int_{t_n}^{t_{n+1}} \mathbf{f}(\x^\varepsilon(s),\y^\varepsilon(s)) ds + \int_{t_n}^{t_{n+1}} \boldsymbol{\sigma}(\x^\varepsilon(s),\y^\varepsilon(s)) d\mathbf{W}_n^1(s).
\ee
For the multiscale system with $\varepsilon \ll 1$, there exists a time scale $\tau\sim O(\varepsilon)$ such that for $t>\tau$ the fast variable $\y$ becomes slaved to the slow variable $\x$, i.e.,
\be \label{fast_equ}
    \y^\varepsilon(t) \approx \y(\x^\varepsilon(t)) \sim \mathcal{P}_{\x}, \qquad t> \tau,
\ee
where $\mathcal{P}_{\x}$ is its (conditioned) invariant measure that is a function of $\x$ only.
Therefore, on any time scale larger than $\tau\sim O(\varepsilon)$, there exists an unknown mapping $\G$ such that
\be \label{sFM}
    \x^\varepsilon_{n+1} \approx \G (\x^\varepsilon_n, d\mathbf{W}_n(\Delta)),
\ee
where $d\mathbf{W}_n(s) = (d\mathbf{W}^1(t_n +s); d\mathbf{W}^2(t_n+s))$
is the independent increment of the Wiener process.
The above argument is illustrated in Figure \ref{fig:multiscale}.

\subsection{Stochastic Flow Map Learning}

Since the slow variables $\x$ follow the unknown mapping \eqref{sFM}, we now seek to construct a stochastic flow map learning (sFML) model in the form of a generative model
\be \label{sFM_model}
\xh_{n+1} = \Gh (\xh_n, \z_n),
\ee
where $\z_n\in\Rs^{n_z}$ is a random variable of known
distribution, e.g., a standard Gaussian, $n_z$ is the dimension of the random input. Our objective is to ensure this is a weak approximation, in distribution, to the unknown mapping \eqref{sFM}. That is, given an exact slow variable $\x^\varepsilon_n$, the sFML model shall produce an approximation of the slow variable $t_{n+1}$,
\be \label{sFML_weak}
\widetilde{\x}^\varepsilon_{n+1} = \Gh (\x^\varepsilon_n, \z_n) \stackrel{d}{\approx} \x^\varepsilon_{n+1}.
\ee
Note that the approximation is in distribution, a weak form of approximation.

\subsubsection{Learning Method and Training Data}

In order to construct the sFML model \eqref{sFM_model} satisfying \eqref{sFML_weak}, we execute the sFML
model for one time step over $\Delta$,
\be \label{sFM1}
\xh_{1} = \Gh (\xh_0, \z_0),
\ee
and utilize the data set \eqref{sqdata} to learn the unknown
operator $\Gh$. 
%

To construct the training data, we reorganize the trajectory data \eqref{sqdata} into pairs, separated by the constant time step $\Delta$,
\be
\left(\x^\varepsilon(t_{k-1}^{(i)}), \x^\varepsilon(t_{k}^{(i)})\right), \qquad \,k=1,...L,\quad i=1,\dots, N_T,
\ee
which contain a total of $M=N_T L$ such data pairs. We rename each pair using indices 0 and 1, and
re-organize the training data set into a set of 
such pairwise data entries, 
\be \label{dataset}
\left(\x_0^{\varepsilon\,(i)}, \x_{1}^{\varepsilon\,(i)}\right), \qquad i=1,\dots, M.
\ee
In other words, the training data set is comprised of $M$ numbers
of very short
trajectories of only two entries. Each of the $i$-th trajectory,
$i=1,\dots, M$,
starts with an ``initial condition'' $\x_0^{\varepsilon\,(i)}$ and ends a single
time step $\Delta$ later at $\x_{1}^{\varepsilon\,(i)}$. 
The one-step sFML model \eqref{sFM1} is then applied to this training data set to accomplish
\be \label{sFML_train}
\widetilde{\x}^\varepsilon_{1} = \Gh (\x^\varepsilon_0, \z_0) \stackrel{d}{\approx} \x^\varepsilon_{1}.
\ee
The presence of the random variable $\z_0$
enables \eqref{sFM1} to be a generative model that can produce random
realizations. Several methods exist to construct stochastic
generative models, e.g., GANs, diffusion model, normalizing flow,
autoencoder, etc. In this paper, we adopt normalizing flow for \eqref{sFM1}.

\subsubsection{Normalizing Flow Model}

Normalizing flows are generative models that can produce efficient and accurate sampling and density evaluation. A normalizing flow is designed to transform a simple probability distribution, e.g., a standard Gaussian, into a more complex distribution by a sequence of diffeomorphisms.
Let $\Z\in\Rs^D$ be a random variable with a known and
tractable distribution $p_\Z$. Let $\g$ be a diffeomorphism, whose
inverse is $\f=\g^{-1}$, and $\Y=\g(\Z)$. By using the change of
variable formula, one obtains the probability of $\Y$:
$$
p_\Y(\y)= p_\Z(\f(\y)) |\det \D\f(\y)|= p_\Z(\f(\y)) |\det \D\g(f(\y))|^{-1},
$$
where $\D\f(\y) = \partial\f/\partial \y$ is the Jacobian of $\f$ and
$\D\g(\z) = \partial \g/\partial \z$ is the Jacobian of $\g$. When the
target complex distribution $p_\Y$ is given, usually as a set of
samples of $\Y$, one chooses to find $\g$ from a parameterized family
$\g_\theta$, where the parameter $\theta$ is optimized to match the
target distribution. Also,
to circumvent the difficulty of constructing a complicated nonlinear
function $\g$, one utilizes a composition of (much) simpler
diffeomorphisms: $\g = \g_m\circ \g_{m-1}\circ\cdots\circ\g_1$. It can
be shown that $\g$ remains a diffeomorphism with its inverse $\f =
\f_1\circ\cdots\circ\f_{m-1}\circ\f_m$. There exists a large amount of
literature on normalizing flows. We refer interested readers to review
articles such as \cite{Kobyzev2021}.


In our setting, we seek to construct the one-step generative model
\eqref{sFM1} by using the training data \eqref{dataset} to accomplish \eqref{sFML_train}. 
We choose $\z_0$ to be a $\ell$-dimensional standard Gaussian since the slow variables are in $\ell$ dimension.
Let $\T_\theta$ be a diffeomorphism with a set of
parameters $\theta\in\Rs^{n_\theta}$. Our objective is to find $\theta$ such that
$\T_\theta(\z_0)$ follows the distribution of $\x_1^\varepsilon$ given by the samples $\{\x_{1}^{\varepsilon\,(i)}\}_{i=1}^M$
in \eqref{dataset}.

Since the distribution of $\x_1^\varepsilon$ clearly depends on the unknown dynamics starting at $\x_0^\varepsilon$, we constraint the choice of $\theta$ to be a function of $\x_0^\varepsilon$, i.e., 
\be \label{theta}
\theta = \N(\x_0^\varepsilon; \Theta),
\ee
where $\N$ is a DNN mapping with trainable hyper-parameters
$\Theta$. This effectively defines
\be \label{x1}
\x_1^\varepsilon = \T_{\N(\x_0^\varepsilon; \Theta)}(\z_0),
\ee
where the diffeomorphism $\T$ is parameterized by the trainable
hyper-parameters $\Theta$ of the DNN. Let  $\mathbf{S}=\T^{-1}$ be the
inverse of $\T$. We have $\z_0 =
\mathbf{S}_{\N(\x_0^\varepsilon ; \mathbf{\Theta})} (\x_1^\varepsilon)$. 
The invertibility of $\T$ allows us to compute:
\be \label{equ:lh}
    p(\x_1^\varepsilon |
    \x_0^\varepsilon;\mathbf{\Theta})=p_{\z_0}\left(\mathbf{S}_{\N(\x_0^\varepsilon; \Theta)}(\x_1^\varepsilon)\right)\left|\det \D{\T_{\N(\x_0^\varepsilon;
          {\Theta})}}(\mathbf{S}_{\N(\x_0^\varepsilon; \Theta)} (\x_1^\varepsilon))\right|^{-1}. 
\ee

The hyperparameters $\Theta$ are determined by maximizing the
expected log-likelihood \eqref{equ:lh}, which is accomplished by minimizing its
negative as the loss
function,
$$
   \mathcal{L}({\Theta}):=-\mathbb{E}_{(\x_0^\varepsilon,\x_1^\varepsilon)\sim p_{\mathtt{data}}} \log(p(\x_1^\varepsilon|\x_0^\varepsilon;{\Theta})),
$$
where $p_{\mathtt{data}}$ is the distribution from the training data
set \eqref{dataset} and computed as
\be \label{loss}
    \mathcal{L}({\Theta}) = -\sum_{j=1}^M \log\left(p(\x_1^{\varepsilon\,(j)} | \x_0^{\varepsilon\,(j)};{\Theta})\right).
    \ee

Several designs for the invertible map $\T$ have been developed and
studied extensively in the literature. These include, for example, masked
autoregressive flow (MAF) \cite{NIPS2017_6c1da886}, real-valued
non-volume preserving (RealNVP) \cite{dinh2017density}, neural
ordinary differential equations (Neural ODE)
\cite{NEURIPS2018_69386f6b}, etc.
In this paper, we adopt the MAF approach, where the dimension of the
parameter $\theta$ in \eqref{theta} is set to be $n_\theta = 2\ell$,
where $\ell$ is the dimension of the slow variables. For the technical
detail of MAF, see \cite{NIPS2017_6c1da886}.

\subsection{DNN Model Structure and System Prediction}

\begin{figure}[htbp]
  \centering
  \includegraphics[width=.8\textwidth]{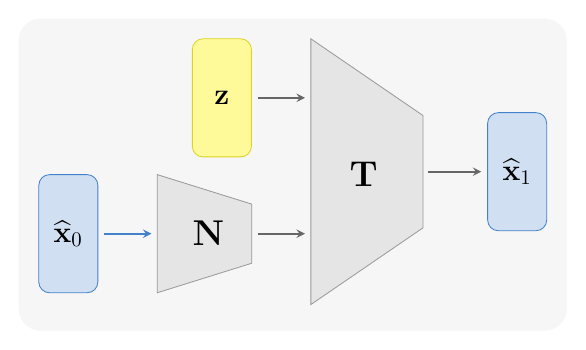}
  \caption{The DNN model structure for the proposed normalizing flow
    sFML method \eqref{x1}.}
  \label{fig:NFnet}
\end{figure}

An illustration of the proposed sFML model structure can be found in Figure
\ref{fig:NFnet}. This is in direct correspondence of \eqref{x1}.
Minimization of the loss function \eqref{loss}, using the data set
\eqref{dataset}, results in the training of the DNN hyperparameters
$\Theta$. Once the training is completed and $\Theta$ fixed,
\eqref{x1} effectively defines the one-step sFML model \eqref{sFM1}:
$$
\x_1 = \T_{\N(\x_0)}(\z_0) = \Gh (\x_0, \z_0),
$$
where we have suppressed the fixed parameter $\Theta$.

The sFML system then produces the system prediction iteratively, for
a given initial condition of the slow variable $\x_0^\varepsilon$,
\be \label{predict}
\left\{
\begin{split}
&\xh_0^\varepsilon = \x_0^\varepsilon, \\
   & \xh_{n+1}^\varepsilon = \Gh (\xh_n^\varepsilon, \z_n), \qquad n\geq 0,
\end{split}
\right.
\ee
where $\z_n$ are i.i.d. $\ell$-dimensional standard normal random
variables. Courtesy of the construction requirement \eqref{sFML_weak}, the sFML prediction is expected to be a weak approximation in distribution to the true stochastic dynamics, i.e., $\xh_n^\varepsilon \stackrel{d}{\approx} \x_n^\varepsilon$, $n\geq 0$. We remark again that the sFML model \eqref{predict} requires only the slow variables. The fast variables are not required during both the training and the prediction of the model. Therefore, the sFML model \eqref{predict} is an effective model for the evolution of the slow variables of the original (unknown) multiscale stochastic dynamical system \eqref{full}.
\section{Numerical Examples}
\label{sec:examples}
In this section, we present several numerical tests to demonstrate the performance of our proposed method. The examples include a skew product SDE, an exponential mean OU process, followed by 3 3-dimensional nonlinear SDEs. In all the examples, the true SDE systems are known. However, the known SDEs are used only to generate the training data set \eqref{dataset}. We solve the true systems by Euler-Maruyama method with a time step $10^{-4}$. The ``initial conditions'' $\x_0$ in \eqref{dataset} are sampled uniformly in a given domain, specified in each example.

In our sFML model, Figure \ref{fig:NFnet},  the DNN has 3 layers, each of which with 20 nodes, and utilizes $\tanh$ activation function. We employ cyclic learning rate with a base rate $3\times10^{-4}$ and a maximum rate $5\times10^{-4}$, $\gamma=0.99999$, and step size $10,000$.  The cycle is set for every $40,000$ training epochs and with a decay scale $0.5$. In our examples, the DNN training is usually conducted for $200,000\sim 300,000$ epochs and batch size is taken $30,000\sim 40,000$.

\subsection{A skew product SDE}
\label{s:ex1}
\begin{figure}[htbp]
  \centering
  \label{fig:ex1_ms}
  \includegraphics[width=.52\textwidth]{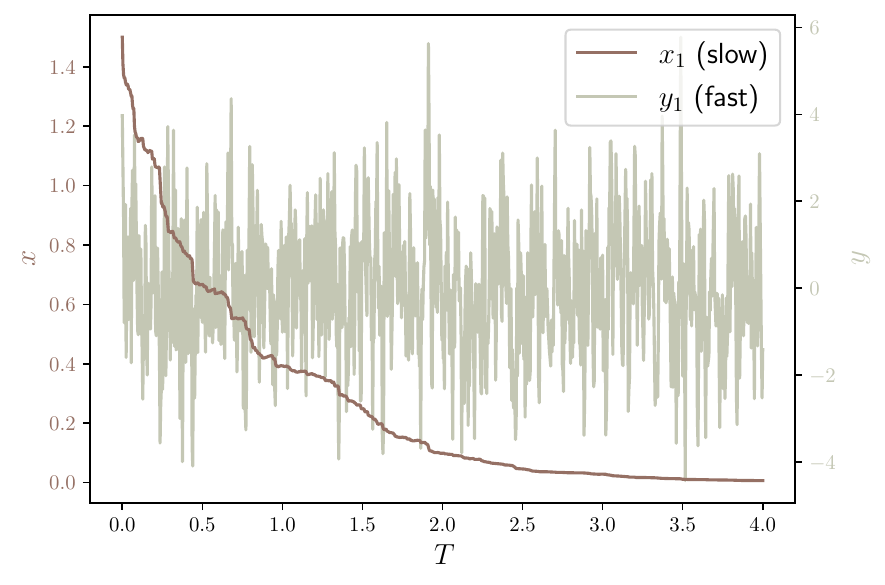}
  \includegraphics[width=.47\textwidth]{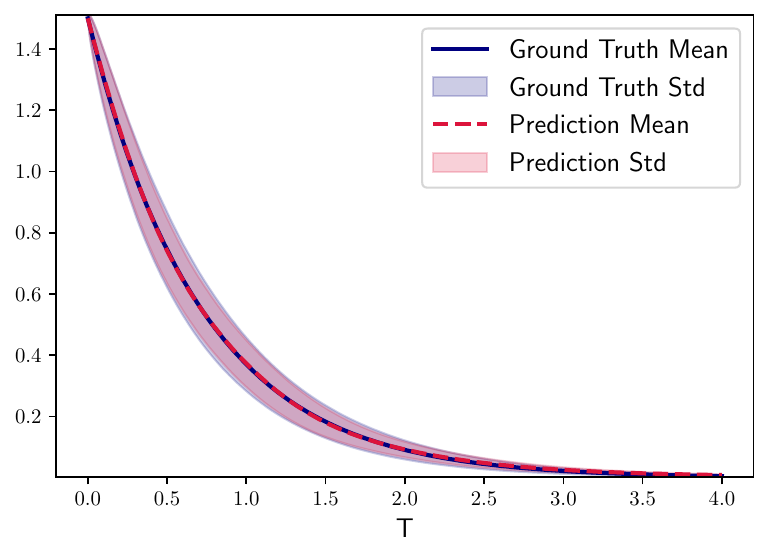}
  \caption{Left: One simulated sample of full system of Example \ref{s:ex1} with initial condition $x_0=1.5$ and $y_0$ sampling from stationary distribution. Right: Comparison for mean and standard deviation (STD) of ground truth and learned sFL model for the slow variable.}
\end{figure}

We first consider the following 2-dimensional stochastic dynamics mentioned in Section 10 of \cite{pavliotis2008multiscale},
\be
\label{ex1}
\left\{
\begin{split}
    & \frac{d x}{d t}=\left(1-y^2\right) x , \\
    & \frac{d y}{d t}=-\frac{\alpha}{\varepsilon} y+\sqrt{\frac{2 \lambda}{\varepsilon}} \frac{d W_t}{d t},
\end{split}
\right.
\ee
where the parameters are set as $\alpha=1.0$, $\lambda=2.4$, and $\varepsilon=0.005$ . The fast variable $y$ is an Ornstein-Uhlenbeck (OU) process, which has a Gaussian stationary distribution with mean $0$ and variance ${\lambda}/{\alpha}$. Notably, this distribution does not depend on the slow variable $x$. The training data set \eqref{dataset} is generated by uniformly sampling $(x_0,y_0)$ in $(-1.5,2)\times(-1,1.6)$ and solved for up to $T=1.0$. A total of $M=40,000$ trajectory pairs are used in the training data set \eqref{dataset}, where the time step $\Delta$ is taken $0.01$. A simulation of the full system is illustrated on the left of Figure \ref{fig:ex1_ms}.
\begin{figure}[htbp]
  \centering
  \label{fig:ex1_sample}
  \includegraphics[width=.48\textwidth]{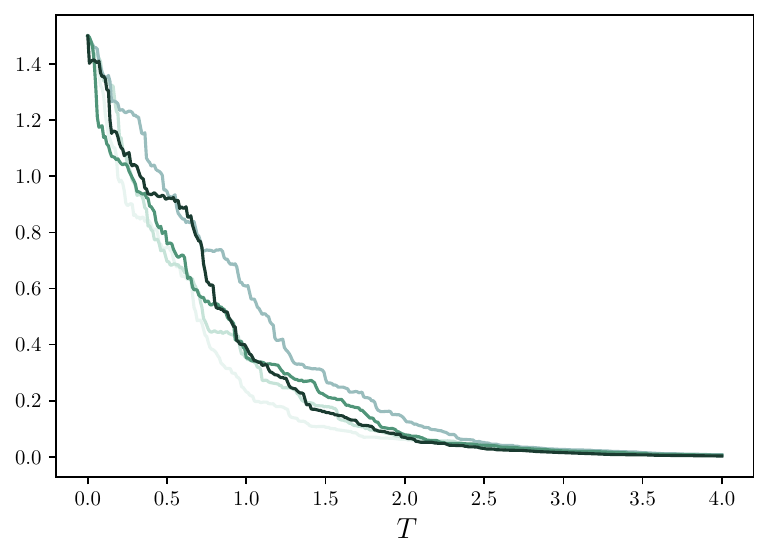}
  \includegraphics[width=.48\textwidth]{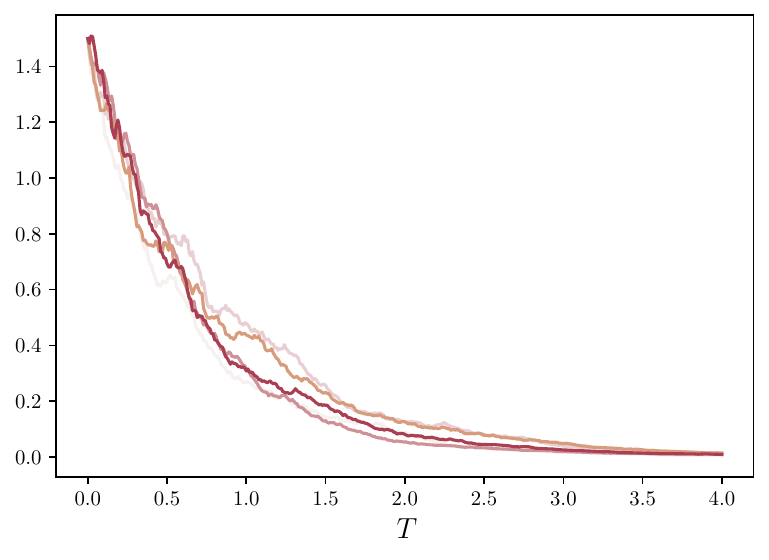}
  \caption{Sample trajectories of Example \ref{s:ex1} slow variable with initial
    condition $x_0=1.5$ and $y_0$ sampling from stationary distribution. Left: ground truth; Right: Simulation using the trained sFML model. }
\end{figure}

Once the sFML model \eqref{sFM_model} is trained, we conduct system predictions up to $T=4.0$, which requires 400 time steps. In Figure \ref{fig:ex1_sample}, we compare some sample trajectory paths produced by the ground truth (left) and the learned sFML model (right), with an initial condition $x_0=1.5$. (For the ground truth, $y_0$ is sampled from the stationary distribution. For the sFML model, $y_0$ is not required.)  We observe the two sets appear visually similar to each other. To further validate the accuracy of sFML method, we compare the mean and standard deviation of the solution averaged over $10,000$ trajectories. The sFML model predictions are shown in the right of Figure \ref{fig:ex1_ms}, along with the reference ground truth. Good agreement is observed.

\subsection{An exponential mean OU process}
\label{s:ex2}

We consider the 2-dimensional stochastic dynamics
\be
\label{ex2}
\left\{
\begin{split}
    & \frac{d x}{d t}=1-x+y, \\
    & \frac{d y}{d t}=\frac{1}{\varepsilon}(e^{-x}-y) +\sqrt{\frac{2}{\varepsilon}} \frac{d W_t}{d t}.
\end{split}
\right.
\ee
Conditioned on the slow variable $x$, the fast variable $y$ is an OU process and eventually converges to a Gaussian stationary distribution with a mean $e^{-x}$ and unit variance. 

\begin{figure}[htbp]
  \centering
  \label{fig:ex2_ms}
  \includegraphics[width=.52\textwidth]{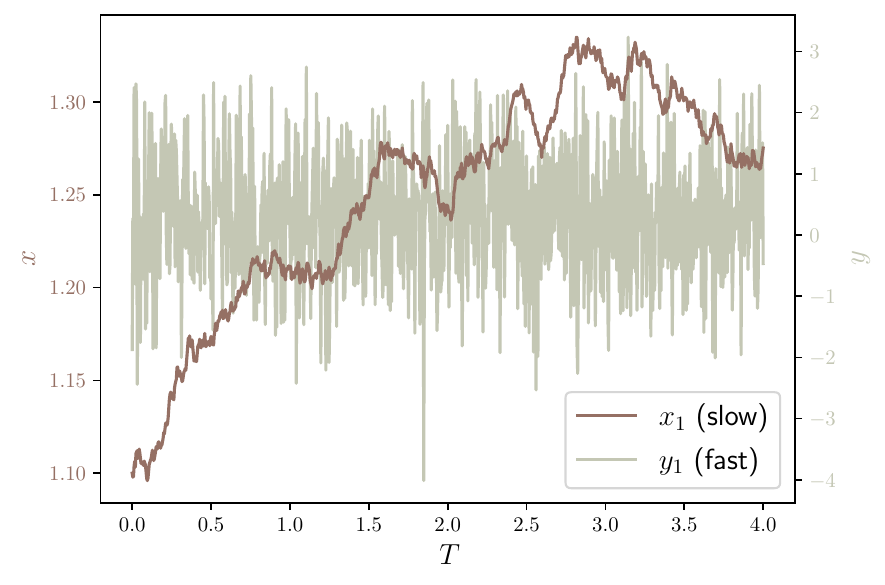}
  \includegraphics[width=.47\textwidth]{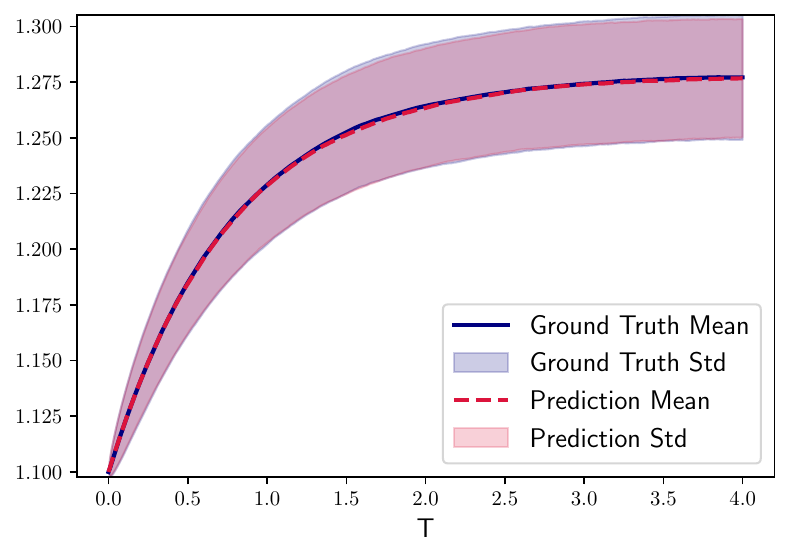}
  \caption{Left: One simulated sample of full system of Example \ref{s:ex2} with initial condition $x_0=1.5$ and $y_0$ sampling from stationary distribution (conditioned on $x_0$). Right: Comparison for mean and standard deviation (STD) of ground truth and learned sFL model for the slow variable.}
\end{figure}

The training data set is generated with $(x_0,y_0)$ uniformly sampled from $(-1.5,2) \times (-4,4)$. We set $\varepsilon=0.001$ and simulate the full system with termination time $T=1.0$ with $\Delta=0.01$. An illustration of the full system can be viewed on the left of Figure \ref{fig:ex2_ms}. A total of $M=40,000$ trajectory pairs are used in the training data set \eqref{dataset}.

\begin{figure}[htbp]
  \centering
  \label{fig:ex2_sample}
  \includegraphics[width=.48\textwidth]{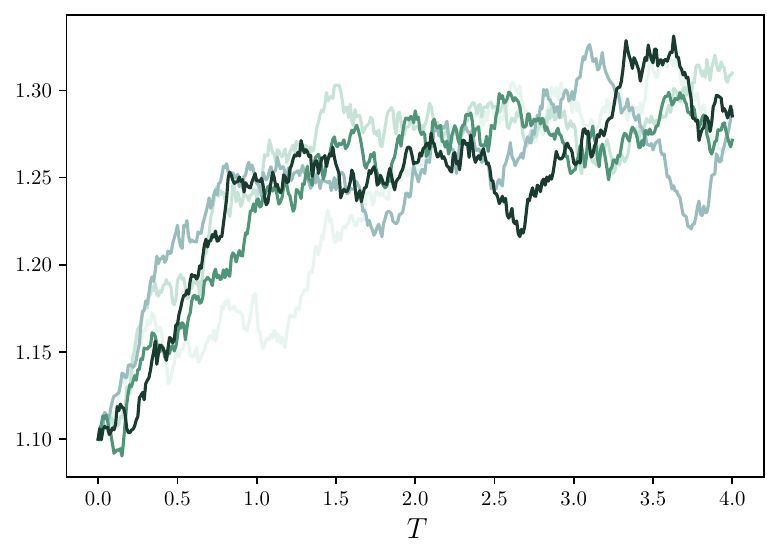}
  \includegraphics[width=.48\textwidth]{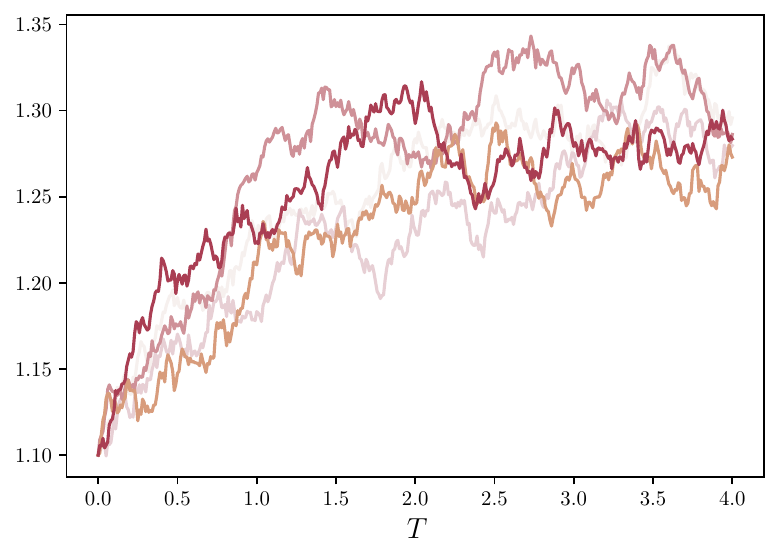}
  \caption{Sample trajectories of Example \ref{s:ex2} slow variable with initial
    condition $x_0=1.5$ and $y_0$ sampling from stationary distribution. Left: ground truth; Right: Simulation using the trained sFML model. }
\end{figure}

With the well-trained sFML model, we compare some sample trajectory paths produced by the ground truth (left) and the learned sFML model (right) in Figure \ref{fig:ex2_sample}. The two sets of samples are visually similar. The mean and STD of the solution are shown in Figure \ref{fig:ex2_ms}. Good agreement is observed for sFML model and reference ground truth.

\subsection{A triad system}
\label{s:ext}
\begin{figure}[htbp]
  \centering
  \label{fig:ext_sim}
  \includegraphics[width=.51\textwidth]{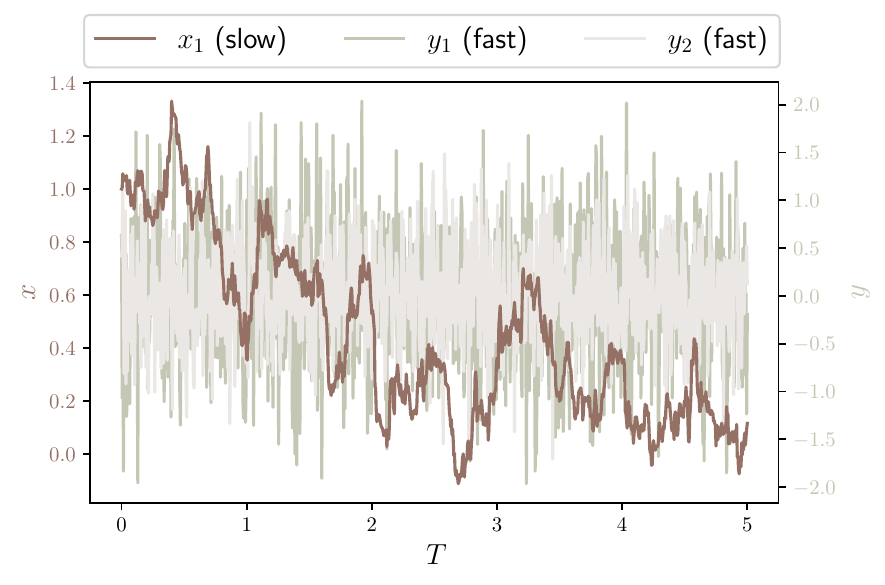}
  \includegraphics[width=.47\textwidth]{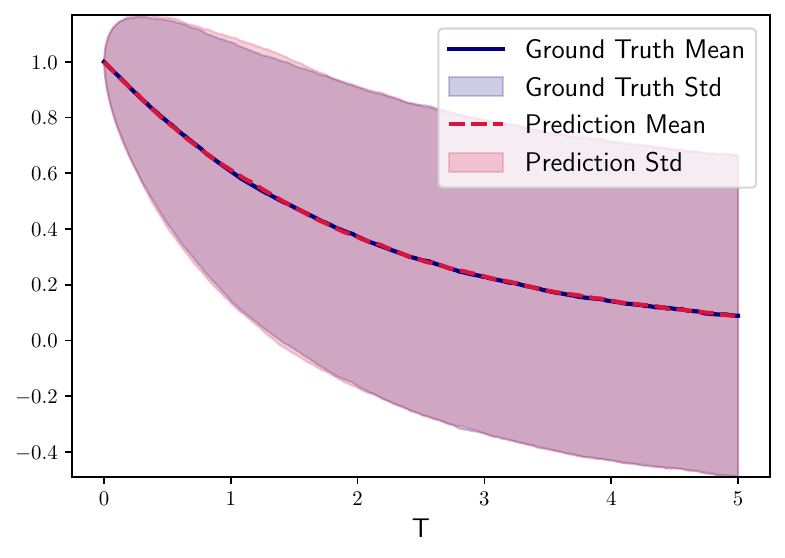}
  \caption{Left: One simulated sample of full system of Example \ref{s:ex3} with initial condition $x_0=1$ and $\y$ sampling from stationary distribution (conditioned on $x_0$). Right: Mean and standard deviation (STD) for slow variable.}
\end{figure}

We consider the following 3-dimensional stochastic dynamics
\be
\label{ext}
\left\{
\begin{split}
    & \frac{d x}{d t}=-\frac{2}{\varepsilon}y_1 y_2, \\
    & \frac{d y_1}{d t}=-\frac{1}{\varepsilon^2}y_1 + \frac{1}{\varepsilon}x y_2 + \frac{\sigma}{\varepsilon} \frac{d W_t^1}{d t},\\
    & \frac{d y_2}{d t}=-\frac{2}{\varepsilon^2}y_2 + \frac{1}{\varepsilon}x y_1 + \frac{\sigma}{\varepsilon} \frac{d W_t^2}{d t},
\end{split}
\right.
\ee
where the parameters are set as $\sigma=1$ and $\varepsilon^2=0.001$. This system is studied as a stochastic model for mean flow-wave interaction in barotropic-baroclinic turbulence \cite{delsole1996quasi, majda1999models}. We are interested in the climate (slow) variable $x$, instead of the weather (fast) variables $\y$. We plot one trajectory sample of the full system on the left of Figure \ref{fig:ext_sim}. According to the work of \cite{MR2165382}, the analytically dervied effective equation for $x$ is $d \overline{x}_t = -\frac{1}{2} \overline{x}_t + \frac{1}{\sqrt{3}}dW_t$, which indicated that the limiting process of $x$ is a OU process. 

Here we utilize the sFML approach to construct a data-driven effective model for $\x$. We generate the training data set up to $T=1$ with initial condition $(x_0,y_0)$ uniformly sampling from $(-2,3)\times (-1,1)^2$ and time step with $\Delta=0.01$. A total $60,000$ trajectory pair data are utilized for training the sFML model.

\begin{figure}[htbp]
  \centering
  \label{fig:ext_sample}
  \includegraphics[width=.48\textwidth]{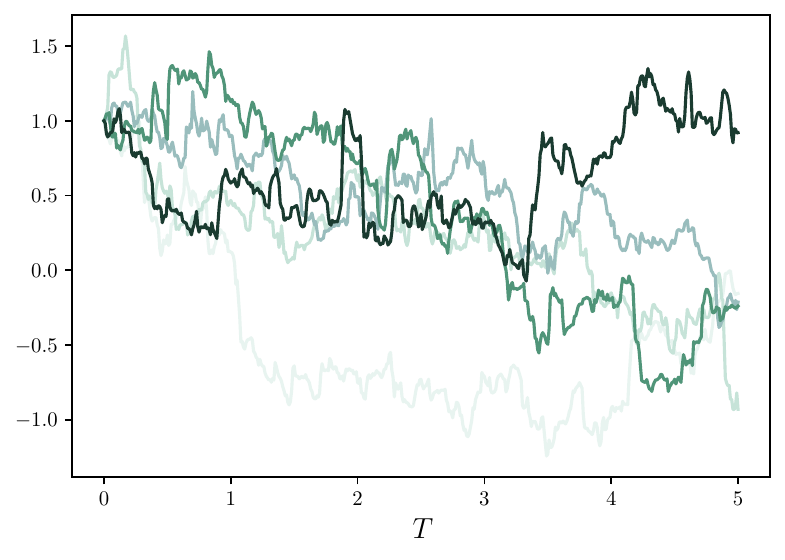}
  \includegraphics[width=.48\textwidth]{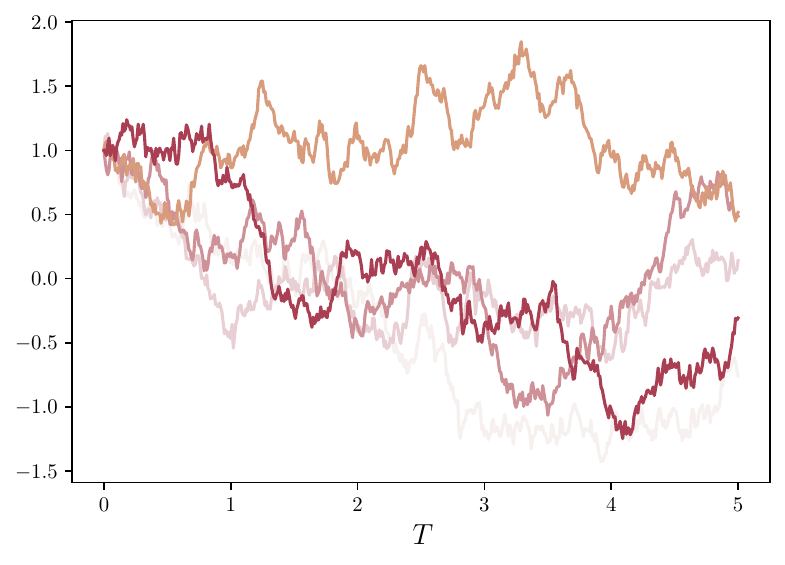}
  \caption{Sample trajectories of Example \ref{s:ext} slow variable with initial
    condition $\x_0=1$ and $\y_0$ sampling from stationary distribution (conditioned on $x_0$). Left: ground truth; Right: Simulation of $x$ using the trained sFML model.}
\end{figure}

Once successfully trained, we generate trajectory samples up to $T=5$ with initial condition $x_0=1$ using the trained model and compare them with ground truth. It takes $500$ steps to generate data from sFML model. Visual comparison of the sample paths for $\x$ is presented in Figure \ref{fig:ext_sample}, where we observe qualitatively similar behavior. To further validate the accuracy, we present the comparisons of mean and STD on the right of Figure \ref{fig:ext_sim}. We observe good agreement between the sFML model prediction and the ground truth. 

\subsection{A 3D nonlinear SDE}
\label{s:ex3}
\begin{figure}[htbp]
  \centering
  \label{fig:ex3_sim}
  \includegraphics[width=.8\textwidth]{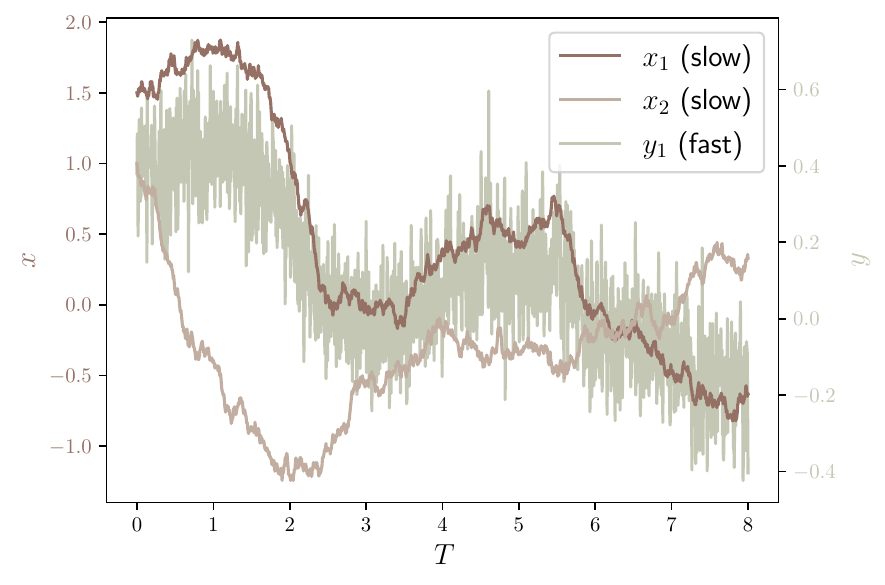}
  \caption{One simulated sample of full system of Example \ref{s:ex3} with initial condition $\x_0=(1.5,1.0)$ and $y_0$ sampling from stationary distribution.}
\end{figure}

We consider the following 3-dimensional stochastic dynamics inspired by \cite{feng2023learning},
\be
\label{ex3}
\left\{
\begin{split}
    & \frac{d x_1}{d t}=x_2 + \sigma_1 \frac{d W_t^1}{d t}, \\
    & \frac{d x_2}{d t}=-x_1-x_2+y^2 + \sigma_2 \frac{d W_t^2}{d t},\\
    & \frac{d y}{d t}=-\frac{1}{\varepsilon} \left( y-\frac{1}{4}x_1\right)+\sigma_3\sqrt{\frac{2}{\varepsilon}} \frac{d W_t^3}{d t},
\end{split}
\right.
\ee
where the parameters are set as $\sigma_1=0.3$, $\sigma_2=0.3$, $\sigma_3=0.1$, and $\varepsilon=0.001$. The one-dimensional fast variable $y$ is driven by an OU process and evolves to a Gaussian distribution with a mean $\frac{1}{4}x_1$ and variance $\sigma_3^2$. 
We plot one trajectory sample of the full system in Figure \ref{fig:ex3_sim} for a visual illustration.

\begin{figure}[htbp]
  \centering
  \label{fig:ex3_sample}
  \includegraphics[width=.48\textwidth]{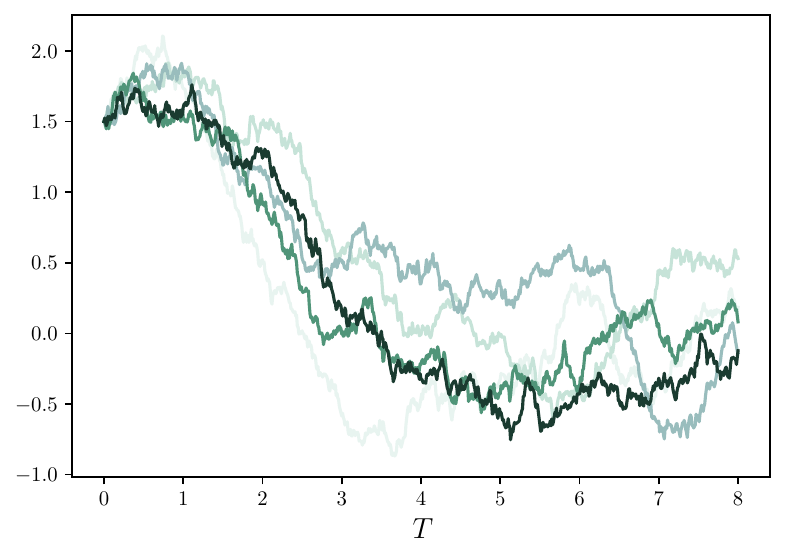}
  \includegraphics[width=.48\textwidth]{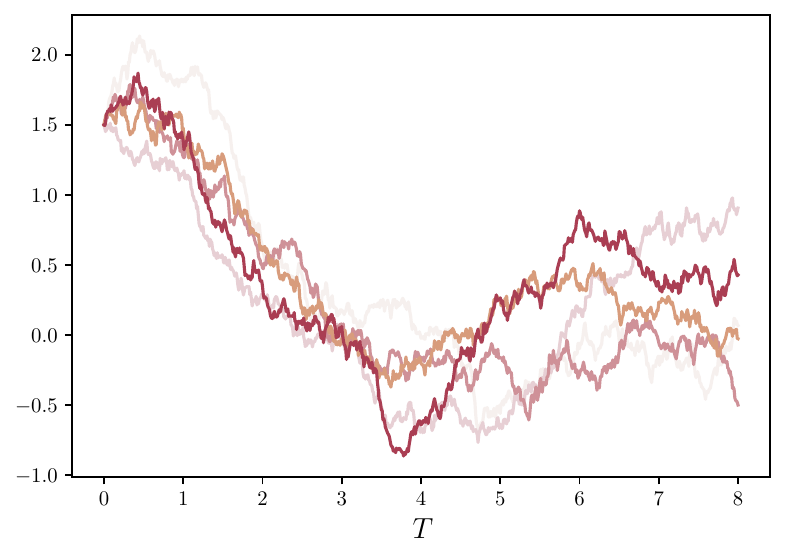}
  \includegraphics[width=.48\textwidth]{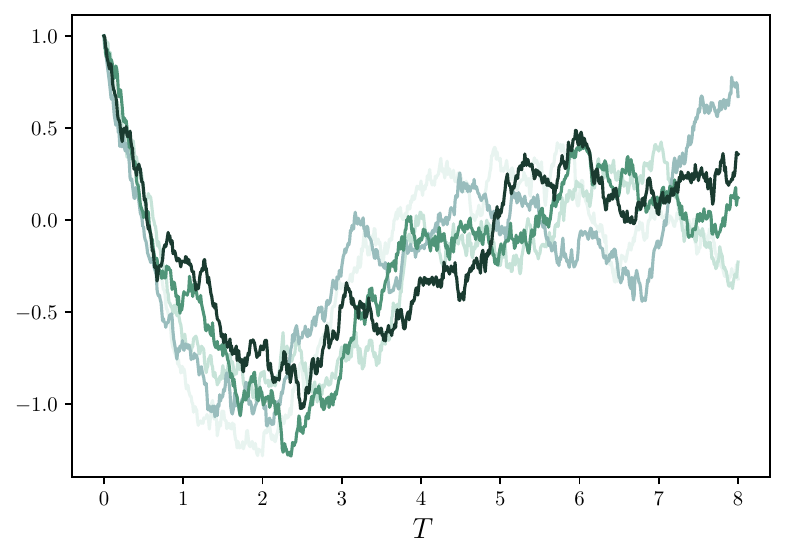}
  \includegraphics[width=.48\textwidth]{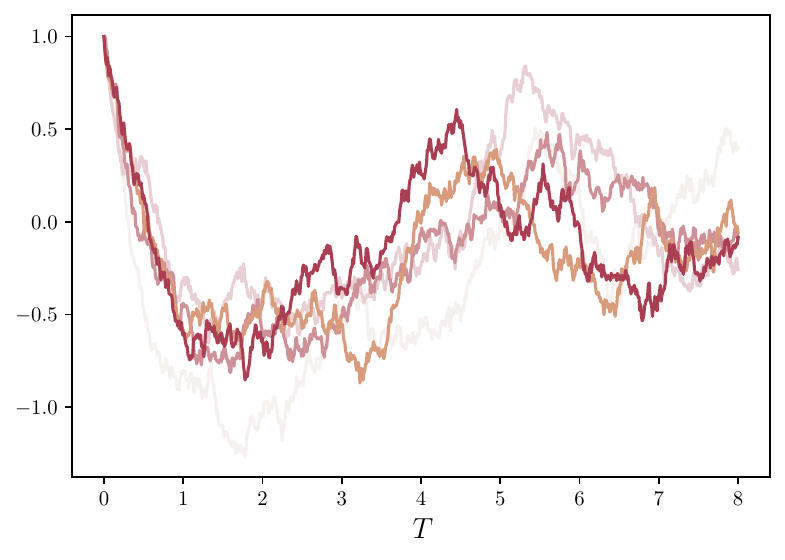}
  \caption{Sample trajectories of Example \ref{s:ex3} slow variable with an initial
    condition $\x_0=(1.5,1.0)$. Left column: ground truth of $x_1$ (top) and $x_2$ (bottom), with $y_0$ sampled from its stationary distribution (conditioned on $\x_0$); Right column: sFML simulation of $x_1$ (top) and $x_2$ (bottom).}
\end{figure}

\begin{figure}[htbp]
  \centering
  \label{fig:ex3_ms}
  \includegraphics[width=.48\textwidth]{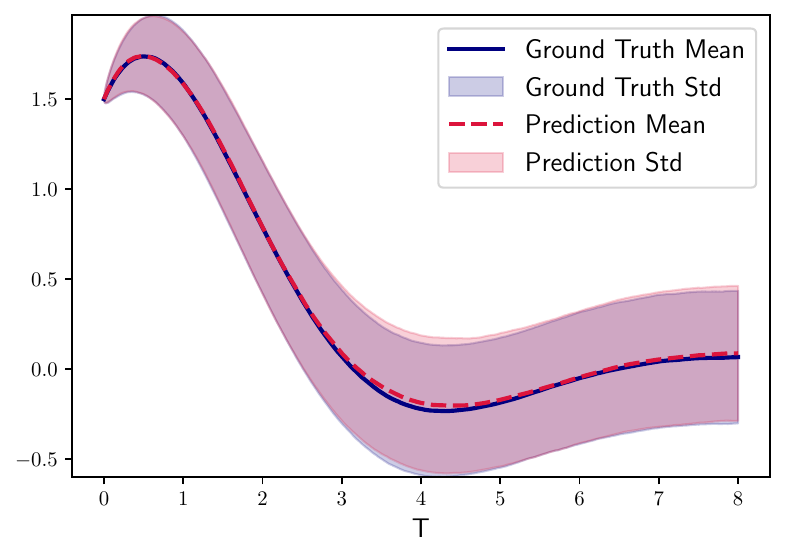}
  \includegraphics[width=.48\textwidth]{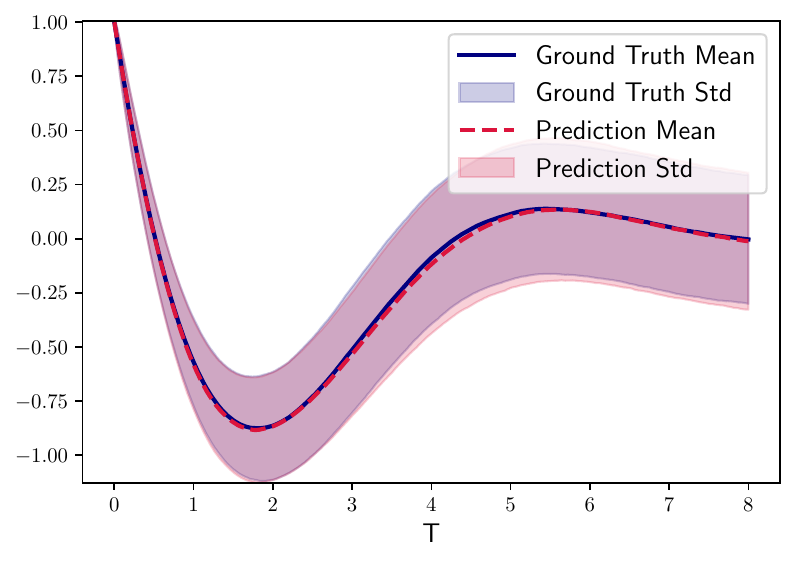}
  \caption{Example \ref{s:ex3}: Mean and standard deviation (STD) for the slow variables $x_1$ (left) and $x_2$ (right), with an initial condition $\x_0=(1.5,1.0)$.}
\end{figure}

\begin{figure}[htbp]
  \centering
  \label{fig:ex3_dis}
  \includegraphics[width=.24\textwidth]{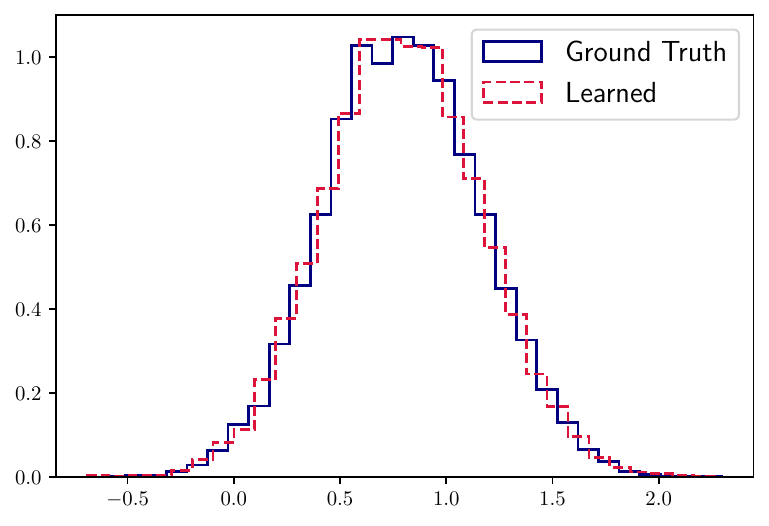}
  \includegraphics[width=.24\textwidth]{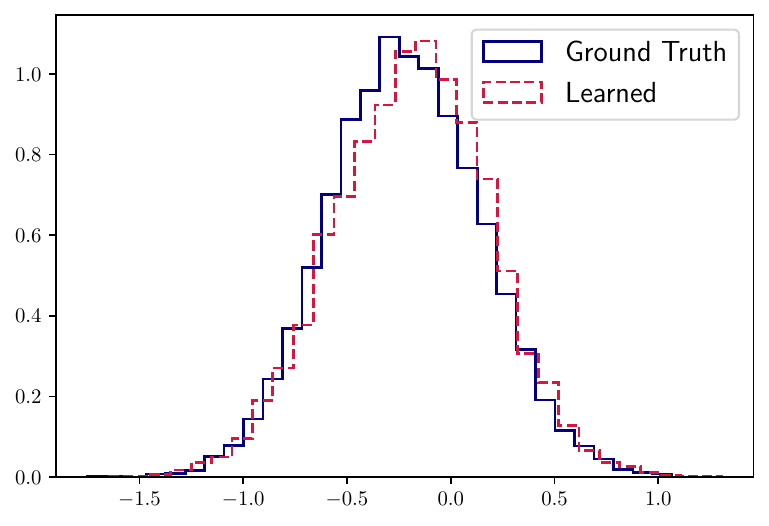}
  \includegraphics[width=.24\textwidth]{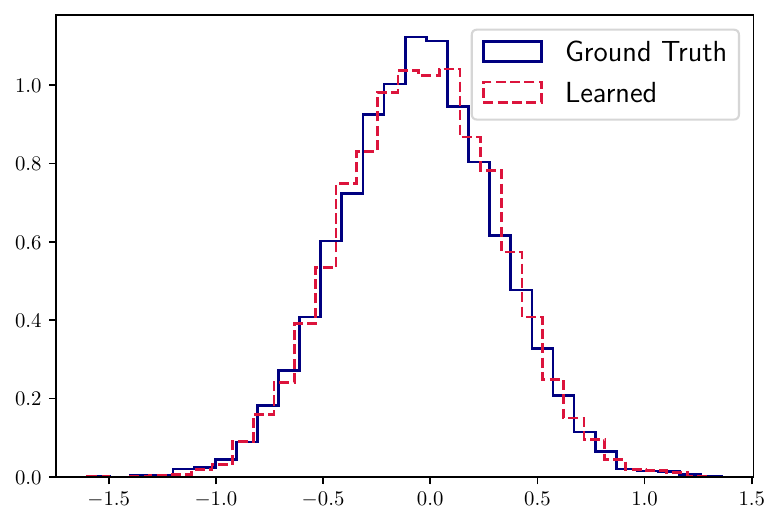}
  \includegraphics[width=.24\textwidth]{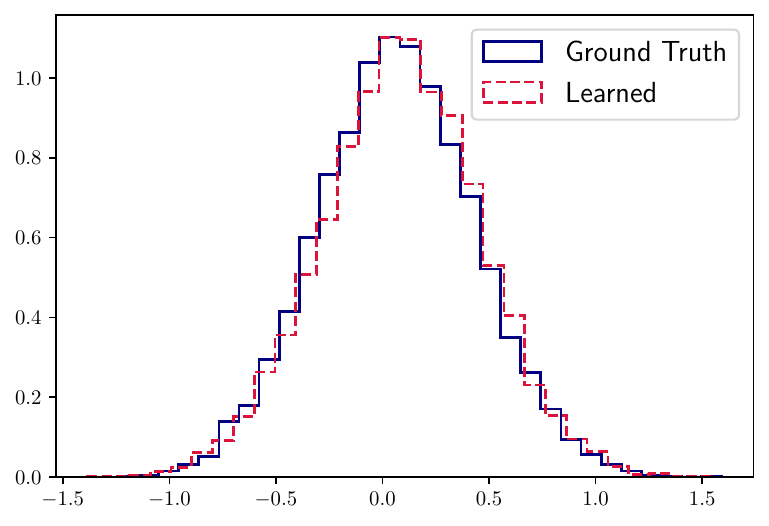}
  \includegraphics[width=.24\textwidth]{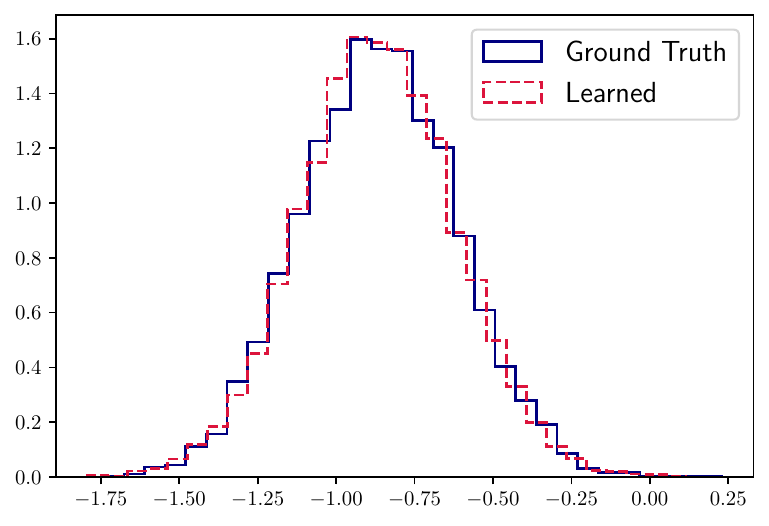}
  \includegraphics[width=.24\textwidth]{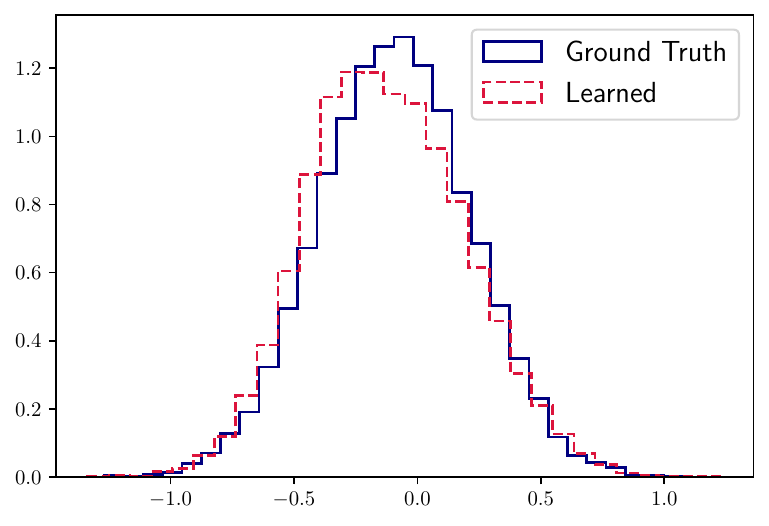}
  \includegraphics[width=.24\textwidth]{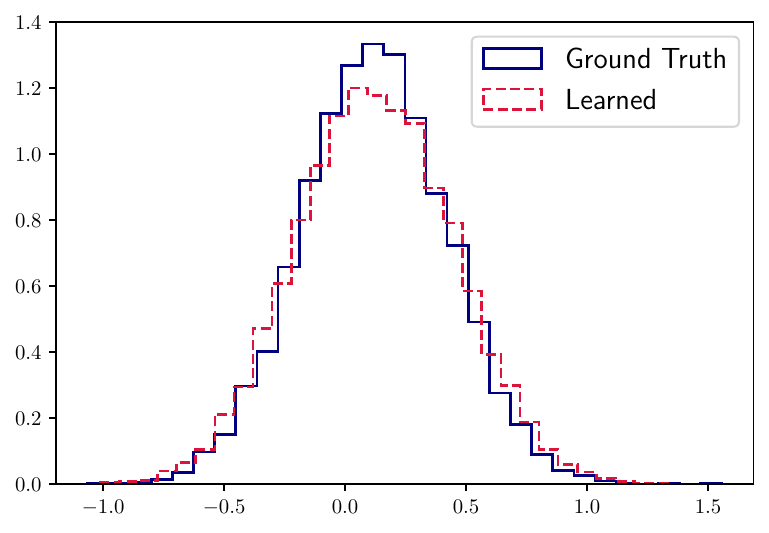}
  \includegraphics[width=.24\textwidth]{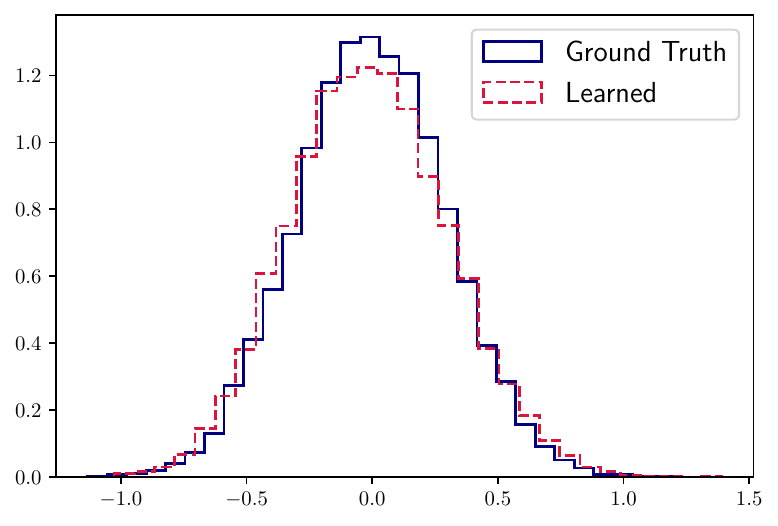}
  \caption{Example \ref{s:ex3}: Comparison of the distribution of the slow variables at $T=2,4,6,8$ with an initial condition $\x_0=(1.5,1.0)$. Top row: $x_1$; Bottom row: $x_2$.}
\end{figure}

To train the sFML effective model,
we generate the training data set up to $T=1$ with initial condition $(\x_0,y_0)$ uniformly sampling from $(-1.5,2.5)\times (-2,1.5) \times (-0.6,1)$ and a time step $\Delta=0.01$. A total of $120,000$ trajectory pair data are utilized for training the sFML model.
Once successfully trained, we generate trajectory samples up to $T=8$ with initial condition $\x_0=(1.5,1.0)$ using the trained sFML model and compare them with ground truth. Visual comparison of the sample paths of the slow variables are shown \ref{fig:ex3_sample}. To further validate the accuracy, we present the comparison of mean and STD in Figure \ref{fig:ex3_ms}. And in
Figure \ref{fig:ex3_dis}, we also show the comparison of the probability distribution of the solution at $T=2, 4,6,8$. We observe good agreement between the sFML model prediction and the ground truth. 

\subsection{Multiscale Stochastic Oscillator}
\label{s:ex4}
\begin{figure}[htbp]
  \centering
  \label{fig:ex4_sim}
  \includegraphics[width=.8\textwidth]{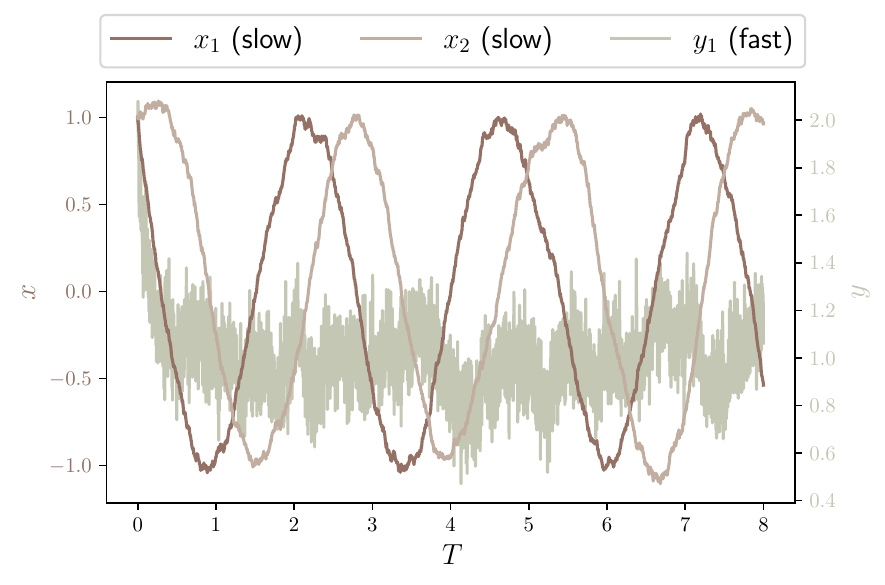}
  \caption{One simulated sample of full system of Example \ref{s:ex4} with initial condition $\x_0=(1.0,1.0)$ and $y_0$ sampling from stationary distribution.}
\end{figure}

We consider the following stochastic oscillator,
\be
\label{ex4}
\left\{
\begin{split}
    & \frac{d x_1}{d t}=\lambda x_1 - \theta x_2 -\gamma x_1 y + \sigma \frac{d W_t^1}{d t}, \\
    & \frac{d x_2}{d t}=\theta x_1 + \lambda x_2 -\gamma x_2 y + \sigma \frac{d W_t^2}{d t},\\
    & \frac{d y}{d t}=-\frac{1}{\varepsilon} \left( y-x_1^2-x_2^2\right)+\sigma\sqrt{\frac{2}{\varepsilon}} \frac{d W_t^3}{d t},
\end{split}
\right.
\ee
where the parameters are set as $\lambda=1.0$, $\theta=1.0$, $\gamma=1.0$, $\sigma=0.1$, and $\varepsilon=0.001$. This system is studied in \cite{chekroun2020ruelle}, whose deterministic version was studied in \cite{noack2003hierarchy} as a low-dimensional reduced model for a flow past a circular cylinder. The one-dimensional fast process $y$ is an OU process and evolves to a Gaussian distribution with mean $(x_1^2+x_2^2)$ and variance $\sigma^2$. 
A trajectory sample path of the full system is shown in Figure \ref{fig:ex4_sim} for demonstration purposes.

\begin{figure}[htbp]
  \centering
  \label{fig:ex4_sample}
  \includegraphics[width=.48\textwidth]{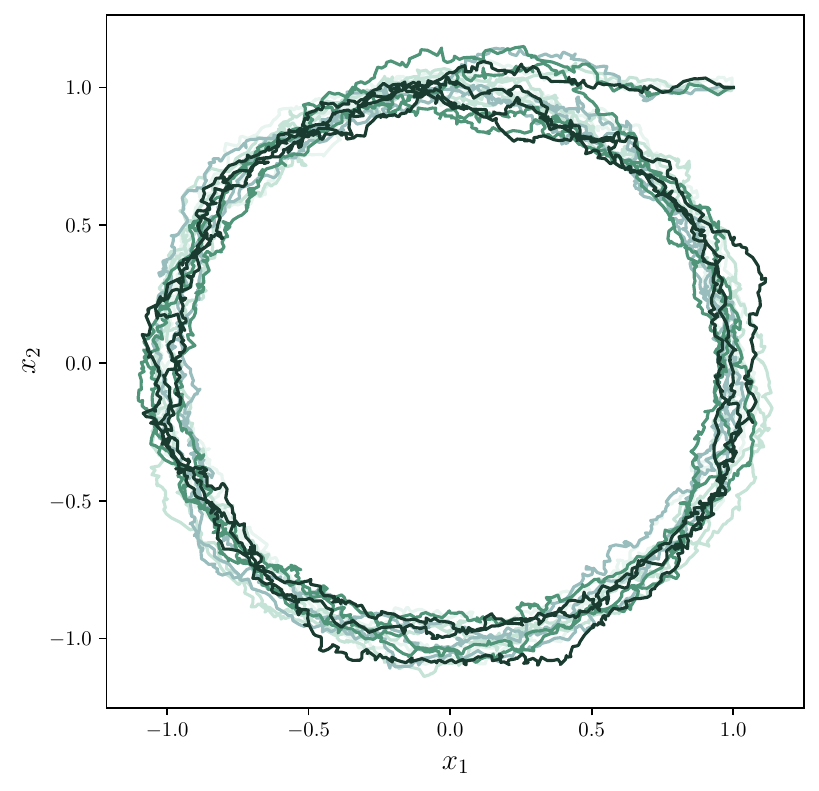}
  \includegraphics[width=.48\textwidth]{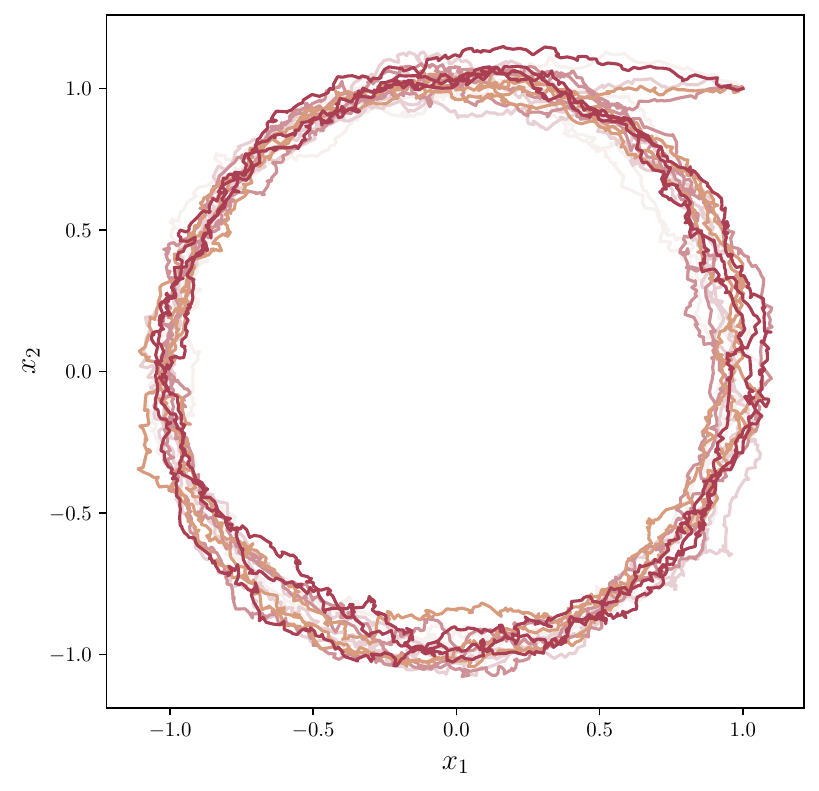}
  \caption{Example \ref{s:ex4}: Samples of phase portrait of the slow variables $(x_1,x_2)$ with initial
    condition $\x_0=(1.0,1.0)$. Left: ground truth (with $y_0$ sampled from its stationary distribution); Right: sFML model.}
\end{figure}

\begin{figure}[htbp]
  \centering
  \label{fig:ex4_ms}
  \includegraphics[width=.48\textwidth]{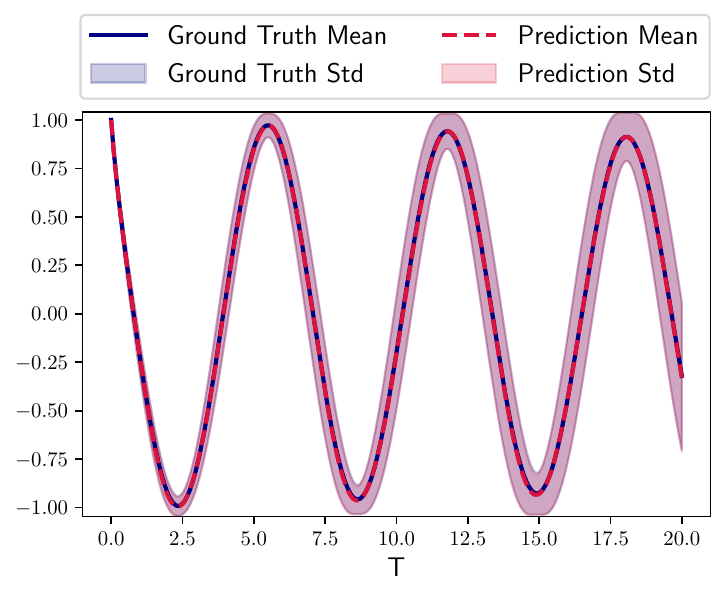}
  \includegraphics[width=.48\textwidth]{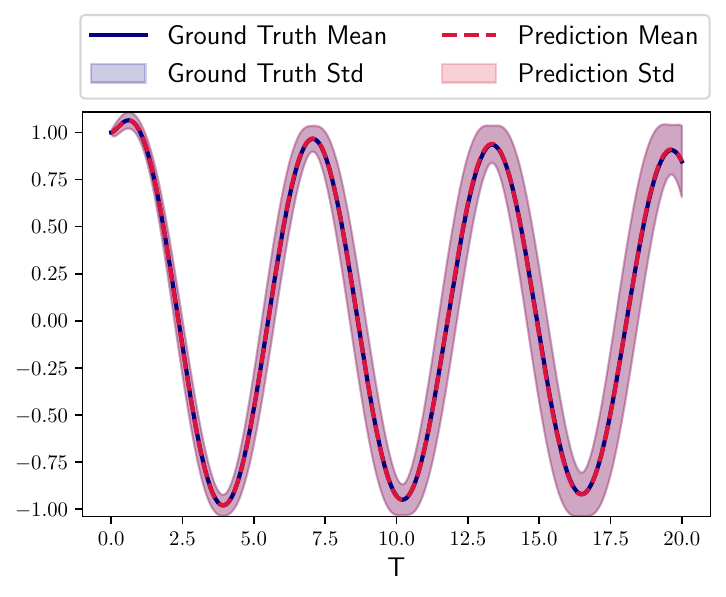}
  \caption{Example \ref{s:ex4}: Comparison of the mean and standard deviation (STD) for slow variables with initial condition $\x_0=(1.0,1.0)$. Left: $x_1$; Right: $x_2$}
\end{figure}

\begin{figure}[htbp]
  \centering
  \label{fig:ex4_dis}
  \includegraphics[width=.24\textwidth]{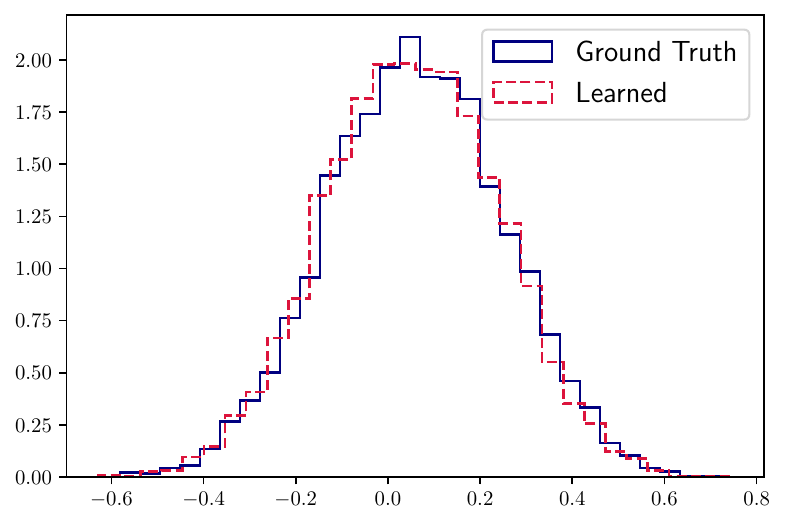}
  \includegraphics[width=.24\textwidth]{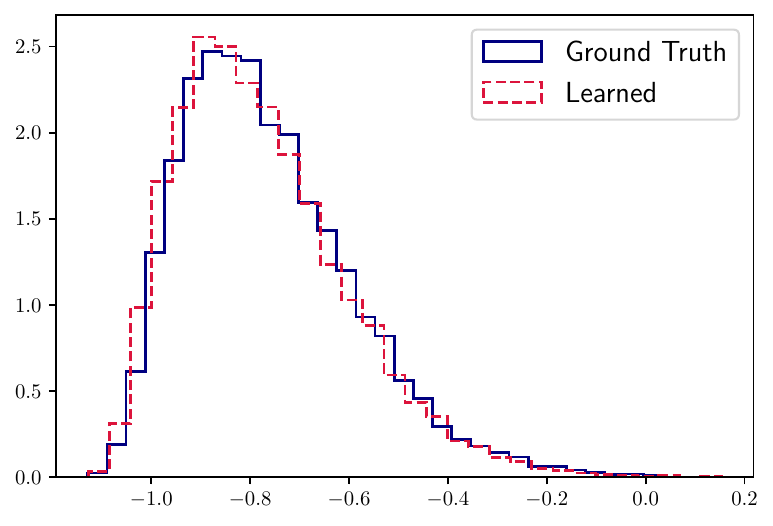}
  \includegraphics[width=.24\textwidth]{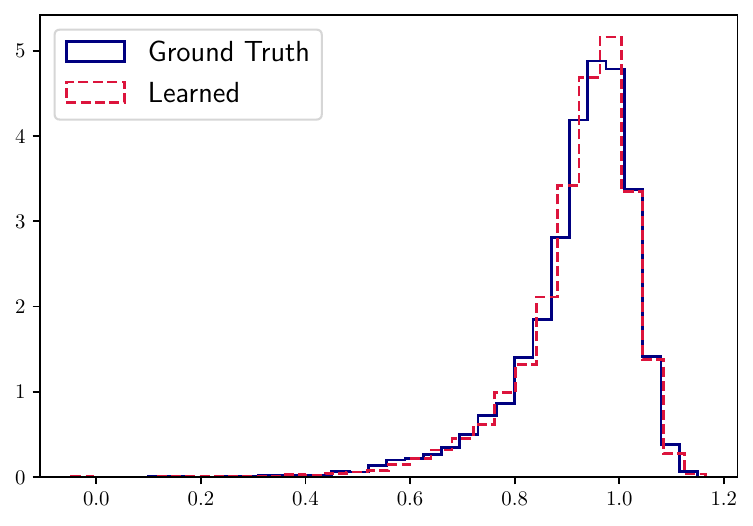}
  \includegraphics[width=.24\textwidth]{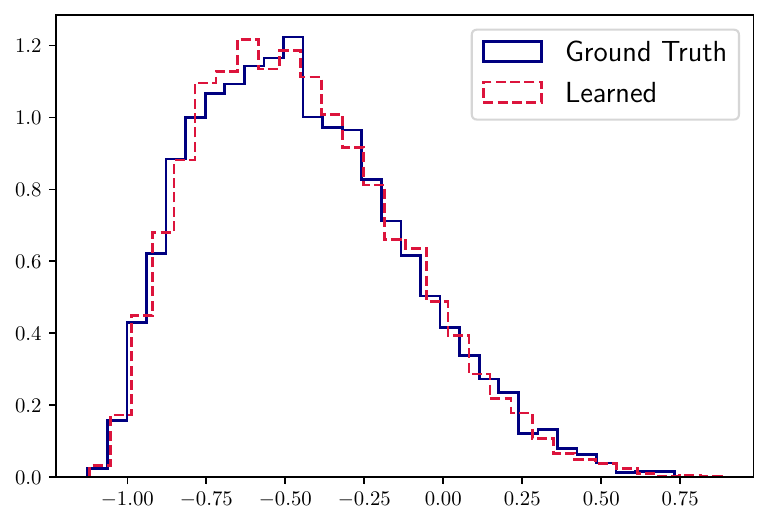}
  \includegraphics[width=.24\textwidth]{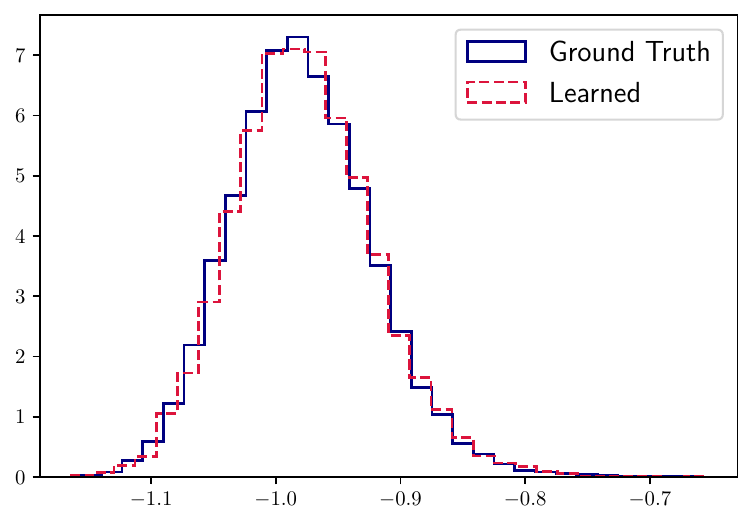}
  \includegraphics[width=.24\textwidth]{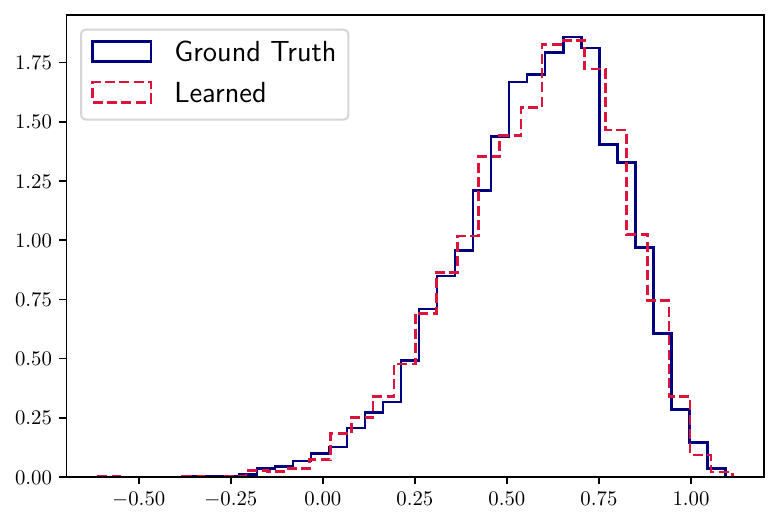}
  \includegraphics[width=.24\textwidth]{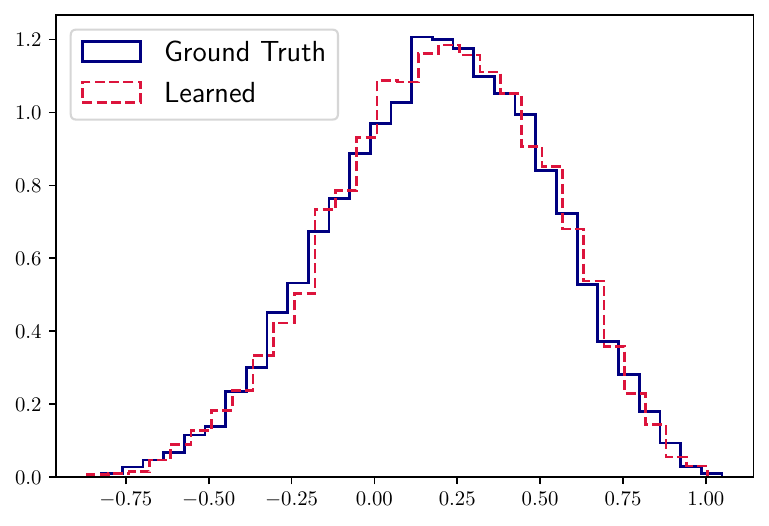}
  \includegraphics[width=.24\textwidth]{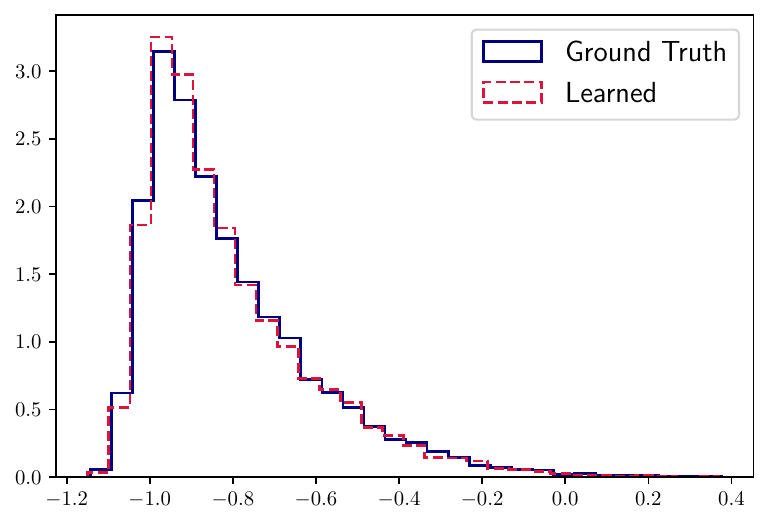}
  \caption{Example \ref{s:ex4}: Comparasion of the distribution of the slow variables at $T=4,8,12,16$ with initial condition $\x_0=(1.0,1.0)$. Top row: $x_1$; Bottom row: $x_2$.}
\end{figure}

To train the sFML effective model, we generate the training data set up to $T=1$ with initial condition $(\x_0,y_0)$ uniformly sampling from $(-1.5,1.5)^2 \times (0.1,2.5)$ and time step $\Delta=0.01$. A total of $120,000$ trajectory data pairs are utilized for training the sFML model.
Once the sFML model is obtained, we generate system prediction for up to $T=20$ with initial condition $\x_0=(1.0,1.0)$ and compare them with ground truth. Samples of the phase portraits of the slow variables are shown in Figure \ref{fig:ex4_sample}, where we observe good visual comparison. To further validate the accuracy, we present the comparison of mean and STD in Figure \ref{fig:ex4_ms}, and the probability distributions at time $T=4, 8,12,16$ in 
Figure \ref{fig:ex4_dis}.  We observe good agreement between the sFML model prediction and the ground truth. 

\section{Conclusions}
In this paper, we presented a numerical method for constructing accurate effective numerical model of slow-fast multiscale stochastic dynamical systems. The method utilizes short bursts of data of the slow variables and constructs the stochastic flow map of the slow dynamics. A conditional normalizing flow model is employed to ensure the learned model is an accurate generative model in distribution. Once successfully trained, the model can serve as a time stepper to produce predictions of the slow variables, with any given initial conditions. By using a comprehensive set of numerical examples, we demonstrated that the proposed approach is effective and accurate in modeling a variety of unknown multiscale stochastic systems.

\bibliographystyle{siamplain}
\bibliography{references}
\end{document}